# A New Learning Paradigm for Stochastic Configuration Network: SCN+

Yanshuang Ao [a], Xinyu Zhou [a], Wei Dai [a,b,*]

[a] School of Information and Control Engineering, China University of Mining and Technology, Xuzhou 221116, China;
[b] State Key Laboratory of Synthetical Automation for Process Industries, Northeastern University, Shenyang 110819, China

*Keywords:*
Learning using privileged information
Stochastic configuration networks
Randomized learner model
Incremental learning
Constructive method

ABSTRACT

Learning using privileged information (LUPI) paradigm, which pioneered teacher-student interaction mechanism, makes the learning models use additional information in training stage. This paper is the first to propose an incremental learning algorithm with LUPI paradigm for stochastic configuration network (SCN), named SCN+. This novel algorithm can leverage privileged information into SCN in the training stage, which provides a new method to train SCN. Moreover, the convergences have been studied in this paper. Finally, experimental results indicate that SCN+ indeed performs favorably.

## 1. Introduction

Gradient algorithms [1] are currently the most widely used in neural network learnings. Although an increasing number of works are aiming at further improving the performance of gradient algorithms, there are still some problems, such as slow convergence, easy to fall into local minimum, strong dependence on the setting of initial parameters, etc.[2]. Thus, it is meaningful to establish a strategy that could make learning process more quickly and more efficient.

Random weight neural networks (RWNNs) are representative examples and have received extensive attention due to its fast learning speed, simplicity of implementation, and excellent generalization performance[3] [4]. In 1994, Pao et al. [5] presented random vector functional link (RVFL) networks, which add a direct link between the input and the output layers in the single layer feedforward neural networks. It should be noted here that RVFL networks calculate the network weights connected to the output layer nodes by solving simple linear regression problems, while other network weights and biases are randomly generated in a specific interval based on a given probability distribution[6] [7]. This type of algorithms does not need to iterate repeatedly, which overcomes the bottleneck problem encountered by traditional gradient algorithms.

Wang et al. [8]proposed a novel random incremental learning technology named stochastic configuration networks (SCNs). By randomly assigning hidden layer parameters within an adjustable interval and innovatively introducing supervisory mechanism to constrain them, SCNs ensure the universal approximation property. Besides, as an incremental learning model, SCNs establish a candidate node pool in each incremental learning process to select the best nodes, which speeds up the convergence rate. At the same time, the network structure could also be determined in the incremental learning process. Aiming at the selection of key parameters in the supervisory mechanism, Zhu et al. [9] gave two kinds of inequality constraints instructively to improve the efficiency of random parameter configuration and help for establishing the candidate node pools. In [10], an ensemble learning method for quickly disassociating heterogeneous neurons was proposed. It used SCNs as base models and adopted a negative correlation learning strategy to evaluate output weights. The TSVD-SCN model was proposed in [11] and aimed at solving ill-posed conditions such as multicollinearity that may exist in the output matrix of the hidden layer. Besides, the truncated singular value decomposition was used to perform generalized inverse operations on the reconstructed output matrix. Wang et al. [12] improved the generalization performance of SCNs by introducing regularization techniques. Wang and Li [13] extended SCNs to deep structure (DeepSCNs) and reported the theoretical analysis and algorithm realization. The results showed that DeepSCNs could be generated quickly and efficiently in a self-organization way, and the generalization performance is similar with the other state-of-the-art deep neural networks. The use of one-dimensional SCNs might destroy the spatial information of the image and lead to poor performance in the tasks of image data modeling. Thus, Li and Wang [14] expanded the original SCNs into a two-dimensional version called 2D-SCNs, which shows good potential for image data analytics. To solve the uncertain data regression problem, two different robust SCNs with kernel density estimation and maximum correntropy criterion were proposed in [15] and [16], respectively. For solving the prediction interval estimation problems that are commonly presented in industrial processes, Lu and Ding et al. [17] developed the corresponding deep ensemble, and robust versions of SCNs.

*Corresponding author.
  *Email address:* weidai@cumt.edu.cn
  This work was supported by the Nature Science Foundation of Jiangsu Province [grant number BK20160275]; the Postdoctoral Science Foundation of China [grant numbers 2015M581885, 2018T110571]; and the Open Project Foundation of State Key Laboratory of Synthetical Automation for Process Industries [grant number PAL-N201706].

The existing algorithms on SCNs have not considered cases that some data are available in the training stage but unavailable in the test stage. Suppose our goal is to determine whether a patient has cancer based on the biopsy image. The problem here is to find the classification rules in the pixel space given the image described in the pixel space. However, in addition to the pictures, the doctor also has a report written by a pathologist, which uses high-level language to describe the pictures. The problem is that if you train the pictures together with the pathologist's report, you will not be able to get the pathologist's report in time when predicting new pictures. In fact, the goal here is to make an accurate diagnosis without consulting a pathologist. However, the pathologist's diagnosis has very important guiding significance for judging the classification of these pictures. Unfortunately, original SCNs is hard to utilize this instructive additional information that is different from the original feature space into the training phase for more accurately predicting.

In order to solve the above problem, learning using privileged information (LUPI) paradigm [18] was first proposed by Vapnik and Vashist. They provided a support vector machine algorithm with privileged information termed SVM+. LUPI paradigm fasten the learning attention on some elements of human teaching. It considers that a teacher who can provide students with explanations plays a significant role in human learning. In contrast to the traditional learning paradigm, LUPI paradigm can provide a set of additional information for the training data during the training process. By employing privileged information, LUPI paradigm solves an open problem of how the data modalities being only available during training time help the learning model achieve better prediction. The new learning paradigm is incorporated into weighted SVM [19], metric learning [20] [21], learning to rank [22], RVFL networks [23] and so on. However, the existing algorithms with LUPI paradigm are all based on batch algorithms, which have certain limitations. There are two essential defects make the application of batch algorithms restricted to some extent: 1) The scale of hidden nodes cannot be determined before training, which means it is difficult to set a suitable network structure; 2) The hidden layer parameters are always generated in a constant interval, which affects its actual approximation property [24] [25].

Moreover, comparing with pruning algorithms, the incremental learning algorithms in SCNs have more advantages. First, the incremental learning algorithms directly define a small initial network, but the pruning algorithms do not know how large the initial network is. Second, incremental learning algorithms always search for small solutions, while pruning algorithms spend most of their training time on networks larger than the optimal solution, which is more time-consuming. Third, there are many different size networks may be acceptable, incremental learning algorithms are likely to find smaller network solutions than pruning algorithms. Finally, the supervisory mechanism adopted by SCNs ensures the universal approximation characteristics of the networks.

In order to enable SCN to use privileged information to build a more generalized model, an incremental learning algorithm and a supervisory mechanism with LUPI paradigm for RVFL is proposed, termed SCN+. The convergence of SCN+ has been proved in this paper, which provides a strong theoretical support. Finally, experiments on the datasets in the machine learning database verify the effectiveness of SCN+ for classification and regression tasks. The work of this paper is mainly carried out from the perspective of algorithm improvement without considering the practical application.

The rest of this paper is as follows: Section 2 briefly describes SCNs and LUPI. Section 3 introduces SCNs+ in detail, including theoretical analysis and algorithm description. Section 4 reports performance evaluation results. Section 5 summarizes this paper and makes some comments on the follow-up work.

## 2. Related work

Given $N$ samples $(x_i, t_i)$, input $X = \{x_1, x_2, \ldots, x_N\}$, $x_i = \{x_{i,1}, \ldots, x_{i,d}\} \in R^d$, corresponding output $T = \{t_1, t_2, \ldots, t_N\}$, $t_i = \{t_{i,1}, \ldots, t_{i,m}\} \in R^m$, with $i = 1, 2, \ldots, N$. SCNs with $L$ hidden nodes can be expressed as:

$$f_L(X) = \sum_{j=1}^{L} H_j(\omega_j, b_j, X)\beta_j, \qquad (1)$$

where $H_j(\cdot)$ represents activation function of the $j$-th hidden node; the hidden-node parameters ($w_j$ and $b_j$) are randomly assigned from $[-\lambda, \lambda]^d$ and $[-\lambda, \lambda]$ respectively; $\beta_j = [\beta_{j,1},\ldots, \beta_{j,q},\ldots, \beta_{j,m}]^T$ expresses the output weights between the $j$-th hidden node and the output nodes; $f_L$ denotes the output function of current network.

SCNs adopt incremental learning method to build model.

2.1. SCNs

In SCNs, the hidden nodes are increased individually. That is to say, starting from randomly generating the first node $g_1 = (w_1, b_1, x)$, SCNs will gradually add nodes to the network [8]. The output $f_L$ can be expressed as a specific combination of the previous network $f_{L-1}$ and the newly added node $g_L$ ($v_L$ and $b_L$):

$$f_L(x) = f_{L-1}(x) + H_L\beta_L, \qquad (2)$$

where the output weight of new node can be obtained based on the following formula:

$$\beta_L = \frac{\langle e_{L-1}, H_L \rangle}{\|H_L\|^2}. \tag{3}$$

And the residual of previous network is

$$e_{L-1} = f - f_{L-1} = [e_{L-1,1}, \ldots, e_{L-1,m}]. \tag{4}$$

The supervisory mechanism used in SCNs is presented in Theorem 1.

**Theorem 1** [11]: Let $\Gamma := \{H_1, H_2, H_3, \ldots\}$ be a set of real-valued function, and span($\Gamma$) denotes a function space spanned by $\Gamma$. Suppose that span($\Gamma$) is dense in $L_2$ space and $\forall g \in \Gamma$, $0 < \|H\| < b_g$ for some $b_g \in R^+$. Given $0 < r < 1$ and a non-negative real number sequence $\{\mu_L\}$, with $\lim_{L \to +\infty} \mu_L = 0$ and $\mu_L \leqslant (1-r)$. For $L = 1, 2, \ldots$, denoted by:

$$\delta_L = (1 - r - \mu_L)\|e_{L-1}\|^2. \tag{5}$$

If the random basis function $g_L$ is generated to satisfy the following inequalities:

$$\langle e_{L-1}, H_L \rangle^2 \geq b_g^2 \delta_L. \tag{6}$$

Then, we have $\lim_{L \to +\infty} \|f - f_L\| = 0$.

The network structure and parameter optimization problems can be solved simultaneously in the above learning process.

**Remark 1**: According to the different solution $\beta_L$ of output weight, three algorithmic implementations of SCNs, namely SC-I, SC-II, and SC-III, are developed. (3) is used in SC-I, and other details can be referred to [11].

2.2. Learning using privileged information

The LUPI paradigm can be described as follows: given a set of triplets $(x_1, \tilde{x}_1, y_1), \ldots, (x_N, \tilde{x}_N, y_N)$, $\tilde{x}_i \in \tilde{X}$, $x_i \in X$ generated according to a fixed but unknown probability measure $P(x, \tilde{x}, y)$ find among a given set of functions $f(x, a)$, $a \in \Lambda$, the function $y = f(x, \tilde{a})$ that guarantees the smallest probability of incorrect classification or regression. Generally speaking, the additional information $\tilde{x} \in \tilde{X}$ belongs to the space $\tilde{X}$ which is different from the space $X$. In other words, this is a new learning paradigm that the additional information is available at the training stage but it is not available for the test set.

# 3. Stochastic configuration networks with privileged information

This section details SCN+. First, we describe the algorithm process. Afterwards, the convergence analysis and its implementation are shown.

3.1. Algorithm description

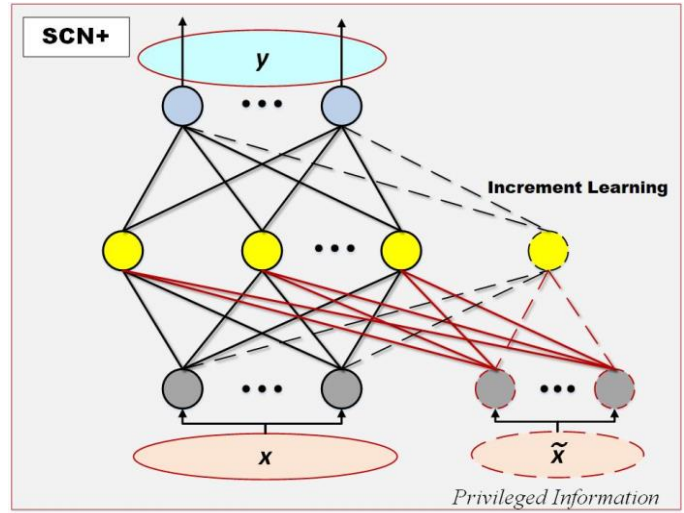

Fig. 1 The structure of the SCN+.

The structure of SCN+ is shown in Fig. 1. In SCN+, the definitions of $x$, $\tilde{x}$ and $y$ are input, additional privileged information and output respectively. Notice that the additional privileged information $\tilde{x}$ is only used in the training stage. Usually, the SCN+ gradually increase the hidden nodes until an acceptable performance is met. The initial residual error $e_0 = f - f_0$, where $f_0 = H_0 \beta_0$, and grown residual error $e_L = f - f_L = f - (f_{L-1} + H_L \beta_L + \tilde{H}_L \tilde{\beta}_L) = e_{L-1} - (H_L \beta_L + \tilde{H}_L \tilde{\beta}_L)$ in the $L$th ($L=1, 2, \ldots, L_{max}$) grown (where

$\beta_L = [\beta_{n+L,1}, \ldots, \beta_{n+m,1}]$ and $\tilde{\beta}_L = [\tilde{\beta}_{n+L,1}, \ldots, \tilde{\beta}_{n+m,1}]$ are output weight vectors of newly added hidden nodes corresponding to the original feature $x$ and privileged feature $\tilde{x}$ respectively in the $L$th grown. The hidden later output vectors

$$H_L = \begin{bmatrix} G(\langle \omega_L, x_1 \rangle + b_L) \\ \vdots \\ G(\langle \omega_L, x_N \rangle + b_L) \end{bmatrix} \text{ and } \tilde{H}_L = \begin{bmatrix} G(\langle \tilde{\omega}_L, \tilde{x}_1 \rangle + \tilde{b}_L) \\ \vdots \\ G(\langle \tilde{\omega}_L, \tilde{x}_N \rangle + \tilde{b}_L) \end{bmatrix}$$

are corresponding to the original feature and privileged feature after the $L$-th grown).

The target function in the constructive process can be formulated as

$$\begin{aligned}
\min_{\Delta \beta_L} f &= \frac{1}{2}\begin{bmatrix} \beta^T & \beta_L^T \end{bmatrix}\begin{bmatrix} \beta \\ \beta_L \end{bmatrix} + \frac{\gamma}{2}\begin{bmatrix} \tilde{\beta}^T & \tilde{\beta}_L^T \end{bmatrix}\begin{bmatrix} \tilde{\beta} \\ \tilde{\beta}_L \end{bmatrix} \\
&+ C\begin{bmatrix} H_L & \tilde{H}_L \end{bmatrix}\begin{bmatrix} \tilde{\beta} \\ \tilde{\beta}_L \end{bmatrix} + \frac{1}{2}\|e_L\| \\
&= \frac{1}{2}\|\beta\|^2 + \frac{1}{2}\|\beta_L\|^2 + \frac{\gamma}{2}\|\tilde{\beta}\| + \frac{\gamma}{2}\|\tilde{\beta}_L\| \\
&+ C\tilde{H}\tilde{\beta} + C\tilde{H}_L\tilde{\beta}_L + \frac{1}{2}\|e_L\|^2 \\
s.t.\ & H\beta + H_L\beta_L = Y - (\tilde{H}\tilde{\beta} + \tilde{H}_L\tilde{\beta}_L)
\end{aligned} \tag{7}$$

where
$\beta = [\beta_1\ \beta_2\ \cdots\ \beta_{L-1}]^T$, $\tilde{\beta} = [\tilde{\beta}_1\ \tilde{\beta}_2\ \cdots\ \tilde{\beta}_{L-1}]^T$,

$$H = \begin{bmatrix} G(\langle \omega_1, x_1 \rangle + b_1) & \cdots & G(\langle \omega_{L-1}, x_1 \rangle + b_{L-1}) \\ \vdots & \ddots & \vdots \\ G(\langle \omega_1, x_N \rangle + b_1) & \cdots & G(\langle \omega_{L-1}, x_N \rangle + b_{L-1}) \end{bmatrix}$$

and

$$\tilde{H} = \begin{bmatrix} G(\langle \tilde{\omega}_1, \tilde{x}_1 \rangle + \tilde{b}_1) & \cdots & G(\langle \tilde{\omega}_{L-1}, \tilde{x}_1 \rangle + \tilde{b}_{L-1}) \\ \vdots & \ddots & \vdots \\ G(\langle \tilde{\omega}_1, \tilde{x}_N \rangle + \tilde{b}_1) & \cdots & G(\langle \tilde{\omega}_{L-1}, \tilde{x}_N \rangle + \tilde{b}_{L-1}) \end{bmatrix}$$

are constructed and fixed before the $L$th grown. $C\tilde{H}\tilde{\beta}$ and $C\tilde{H}_L\tilde{\beta}_L$ are slack functions and $\gamma$ is a regularization coefficient.

Take the partial derivative of the target function with respect to $\beta_L$ and $\tilde{\beta}_L$, we have

$$\frac{\partial f}{\partial \beta_L} = \beta_L + H_L^T(H_L\beta_L + \tilde{H}_L\tilde{\beta}_L - e_{L-1}) = 0 \tag{8}$$

$$\frac{\partial f}{\partial \tilde{\beta}_L} = \gamma\tilde{\beta}_L + \tilde{H}_L^T(H_L\beta_L + \tilde{H}_L\tilde{\beta}_L - e_{L-1}) + C\tilde{H}_L^T l = 0 \tag{9}$$

Combining (1) and (1), the output weights can be calculated as

$$\beta_L = \frac{((\gamma + \tilde{H}_L^T\tilde{H}_L)H_L^T - H_L^T\tilde{H}_L\tilde{H}_L^T)e_{L-1} + CH_L^T\tilde{H}_L\tilde{H}_L^T l}{(1 + H_L^T H_L)(\gamma + \tilde{H}_L^T\tilde{H}_L) - H_L^T\tilde{H}_L\tilde{H}_L^T H_L} \tag{10}$$

$$\tilde{\beta}_L = \frac{(\tilde{H}_L^T H_L H_L^T - (1 + H_L^T H_L)\tilde{H}_L^T)e_{L-1} + C(1 + H_L^T H_L)\tilde{H}_L^T l}{\tilde{H}_L^T H_L H_L^T \tilde{H}_L - (1 + H_L^T H_L)(\gamma + \tilde{H}_L^T\tilde{H}_L)} \tag{11}$$

where $l \in \Re^{N \times m}$ is an identity matrix. Furthermore, we can obtain the output function $f_{test}(z) = H(z)\beta$ in the test stage when using the test data $z$. We call the above algorithm without supervision mechanism IRVFL+.

3.2. Definition of SCN+

**Theorem 2.** Suppose that $span(\Gamma)$ is dense in $L_2$ space. Given $0 < r < 1$ and a nonnegative real number $sequence\{\mu_L\}$, with $\lim_{L \to \infty} \mu_L = 0$ and $\mu_L < (1-r)$. For $L = 1, 2, \ldots$, denoted by

$$\delta_L = (1 - r - \mu_L)\|e_{L-1}\|^2. \tag{12}$$

If the hidden output block vectors $H_L$ and $\tilde{H}_L$ are generated with the following inequalities：

$$\langle e_{L-1}, H_L\beta_L + \tilde{H}_L\tilde{\beta}_L \rangle \geq \delta_L \tag{13}$$

and the output weights $\beta_L$ and $\tilde{\beta}_L$ are evaluated by (10) and (11) respectively, then we have $\lim_{L \to \infty} \|f - f_L\| = 0$.

### 3.3. Convergence analysis

For calculating easily, we reformulate the target function of SCN+ as

$$\min_{\Delta\beta_L} f = \frac{1}{2}\Delta\beta^T A \Delta\beta + \frac{1}{2}\Delta\beta_L^T A \Delta\beta_L + \Delta H_L B \Delta\beta_L$$
$$+ \frac{1}{2}(\Delta H_L \Delta\beta_L - e_{L-1})^T (\Delta H_L \Delta\beta_L - e_{L-1}), \quad (14)$$

where

$$\Delta\beta = \begin{bmatrix} \beta \\ \tilde{\beta} \end{bmatrix}, \Delta\beta_L = \begin{bmatrix} \beta_L \\ \tilde{\beta}_L \end{bmatrix}, A = \begin{bmatrix} 1 & 0 \\ 0 & \gamma \end{bmatrix}, B = \begin{bmatrix} 0 & 0 \\ 0 & C \end{bmatrix},$$
$$\Delta H_L = \begin{bmatrix} H_L & \tilde{H}_L \end{bmatrix}. \quad (15)$$

Take the partial derivative of formula (8) with respect to $\Delta\beta_L$, we have

$$\frac{\partial f}{\partial \Delta\beta_L} = A\Delta\beta_L + B\Delta\tilde{H}_L^T l + \Delta H_L^T (\Delta H_L \Delta\beta_L - e_{L-1}) = 0. \quad (16)$$

Thus, the solution formula for the output weight could be obtained:

$$\Delta\beta_L = (A + \Delta H_L^T \Delta H_L)^\dagger (\Delta H_L^T e_{L-1} - B\Delta H_L^T l). \quad (17)$$

Then, we can prove

$$\|e_L\|^2 - (r + \mu_L)\|e_{L-1}\|^2$$
$$= \langle e_{L-1} - \Delta H_L \Delta\beta_L, e_{L-1} - \Delta H_L \Delta\beta_L \rangle - (r + \mu_L)\langle e_{L-1}, e_{L-1} \rangle$$
$$= (1 - r - \mu_L)\langle e_{L-1}, e_{L-1} \rangle - 2\langle e_{L-1}, \Delta H_L \Delta\beta_L \rangle$$
$$+ \langle \Delta H_L \Delta\beta_L, \Delta H_L \Delta\beta_L \rangle$$
$$= \delta_L - (\Delta H_L^T e_{L-1} - B\Delta H_L^T l)^T ((A + \Delta H_L^T \Delta H_L)^\dagger)^T$$
$$\times (\Delta H_L^T \Delta H_L)^T (A + \Delta H_L^T \Delta H_L)^\dagger (\Delta H_L^T e_{L-1} - B\Delta H_L^T l)$$
$$+ 2e_{L-1}^T \Delta H_L (A + \Delta H_L^T \Delta H_L)^\dagger (\Delta H_L^T e_{L-1} - B\Delta H_L^T l)$$
$$\leq \delta_L - e_{L-1}^T \Delta H_L ((A + \Delta H_L^T \Delta H_L)^\dagger)^T (\Delta H_L^T \Delta H_L)^T$$
$$\times (\Delta H_L^T \Delta H_L)^\dagger (\Delta H_L^T e_{L-1} - B\Delta H_L^T l)$$
$$+ 2e_{L-1}^T \Delta H_L (A + \Delta H_L^T \Delta H_L)^\dagger (\Delta H_L^T e_{L-1} - B\Delta H_L^T l)$$
$$= \delta_L - e_{L-1}^T \Delta H_L (A + \Delta H_L^T \Delta H_L)^\dagger (\Delta H_L^T e_{L-1} - B\Delta H_L^T l)$$
$$= \delta_L - \langle e_{L-1}, \Delta H_L \Delta\beta_L \rangle$$
$$\leq 0 \quad (18)$$

so that we can easily get $\|e_L\| \leq (r + \mu_L)\|e_{L-1}\|$.

Then, suppose $\lim_{L\to\infty} \|e_{L-1}\| = \lim_{L\to\infty} \|e_L\| = a$. Notice that $r + \mu_L < 1$, so the formula $a < (1 + \mu_L)a$ does not hold when $L \to \infty$. Therefore, we have $\lim_{L\to\infty} \|e_{L-1}\| = \lim_{L\to\infty} \|e_L\| = \lim_{L\to\infty} \|f - f_L\| = 0$.

### 4.4. Algorithm implementation

The detailed implementation procedures of SCN+ are summarized as follows.

---
**Algorithm** SCN+

---
**Input**: A set of training data $\{(x_i, \tilde{x}_i, y_i) \mid x_i \in R^n, \tilde{x}_i \in R^d, y_i \in R^m, \forall 1 \leq i \leq N\}$; Set a nonlinear activation function $G(\cdot)$, the maximum number of hidden nodes $L_{\max}$, tolerance error $\varepsilon$, the user-specified coefficients $C$ and $\gamma$, the maximum times of random configuration $T_{\max}$; Choose a set of positive scalars $\Upsilon = \{\lambda_{\min} : \Delta\lambda : \lambda_{\max}\}$;

1. Initialize $e_0 = Y$, $0 < r < 1$, $\Omega = [\ ]$; $W = [\ ]$; $\tilde{W} = [\ ]$;

2. **While** $L \leq L_{\max}$ **AND** $\| e_0 \|_F > \varepsilon$

   **1). Hidden node Parameters Configuration**

3.   **For** $\lambda \in \Upsilon$, **Do**

4.     **For** $i = 1, 2, \ldots, T_{\max}$, **Do**

5.       Randomly assign $\omega_L$, $\tilde{\omega}_L$, $b_L$ and $\tilde{b}_L$ from $[-\lambda, \lambda]^n$, $[-\lambda, \lambda]^d$, $[-\lambda, \lambda]$ and $[-\lambda, \lambda]$;

6.       Calculate $h_L$, $\tilde{h}_L$, $\xi_{L,q}$ based on () and ();

         Set $\mu_L = (1-r)/(L+1)$;

7.       **If** $\min\{\xi_{L,1}, \xi_{L,2}, \ldots, \xi_{L,m}\} \geq 0$

8.         **Save** $\omega_L$ and $b_L$ in $W$, $\tilde{\omega}_L$ and $\tilde{b}_L$ in $\tilde{W}$, $\xi_L = \sum_{q=1}^{m} \xi_{L,q}$ in $\Omega$;

9.       **Else** go back to **Step 4**

10.     **End If**

11.     **End For**

12.     **If** $W$ is not empty

13.       Find $\omega_L^*$, $\tilde{\omega}_L^*$, $b_L^*$ and $\tilde{b}_L^*$ that maximize $\xi_L$ in $\Omega$;

         Set $H_L = [h_1^*, h_2^*, \ldots, h_L^*]$, $\tilde{H}_L = [\tilde{h}_1^*, \tilde{h}_2^*, \ldots, \tilde{h}_L^*]$;

14.       **Break**

15.     **Else** randomly take $\tau \in (0, 1-r)$, renew $r = r + \tau$,

       return to **Step 4**;

16.     **End If**

17.   **End For**

   **2). Output Weights Determination**

18. Calculate $\beta_L$ and $\tilde{\beta}_L$ based on (10) and (11);

   $e_L = e_{L-1} - (H_L \beta_L + \tilde{H}_L \tilde{\beta}_L)$;

19. Renew $e_0 = e_L$; $L = L + 1$;

20. **End While**

21. **Return** $\beta$, $\omega$ and $b$.

## 4. Performance evaluation

### 4.1. Evaluation on real-world datasets

In this section, several experiments are conducted on ten real-world datasets from KEEL. Fifty trials for each method are carried out, and the average results are reported. The experiments are conducted in MATLAB 2019b environment running on a PC that equips with an Intel(R) Core (TM) i7-9700, 3.0 GHz CPU, 8 GB RAM.

Table 1
Statistics of classification datasets.

| Problems | Training data | Testing data | Input | Normal features size | Privileged features size | Output |
|---|---|---|---|---|---|---|
| Mortgage | 700 | 346 | 15 | 8 | 7 | 1 |
| Treasury | 700 | 349 | 15 | 8 | 7 | 1 |
| ENB | 400 | 368 | 8 | 4 | 4 | 1 |
| Laser | 700 | 293 | 4 | 2 | 2 | 1 |
| W-Izmir | 1000 | 461 | 9 | 5 | 4 | 1 |

Table 2
Statistics of regression datasets.

| Problems | Training data | Testing data | Input | Normal features size | Privileged features size | Classes |
|---|---|---|---|---|---|---|
| Wine | 100 | 78 | 13 | 7 | 6 | 3 |
| Contraceptive | 1000 | 473 | 9 | 5 | 4 | 3 |
| Pima | 500 | 268 | 8 | 4 | 4 | 2 |
| Flare | 700 | 366 | 11 | 6 | 5 | 6 |
| ACA | 400 | 290 | 14 | 7 | 7 | 2 |

Table 3
Statistics of parameter settings.

| Data sets | Algorithms parameters ($L_{max}$, $u$, $T_{max}$) | | | |
|---|---|---|---|---|
| | SCN | SCN+ | IRVFL | IRVFL+ |
| Mortgage | 100, {1: 1: 10}, 10 | 100, {1: 1: 10}, 10 | 200, {10}, 1 | 200, {10}, 1 |
| Treasury | 100, {1: 1: 10}, 10 | 100, {1: 1: 10}, 10 | 200, {10}, 1 | 200, {10}, 1 |
| ENB | 100, {1: 1: 10}, 10 | 100, {1: 1: 10}, 10 | 100, {10}, 1 | 100, {10}, 1 |
| Laser | 100, {1: 1: 10}, 10 | 100, {1: 1: 10}, 10 | 100, {10}, 1 | 100, {10}, 1 |
| W-Izmir | 100, {1: 1: 10}, 10 | 100, {1: 1: 10}, 10 | 200, {10}, 1 | 200, {10}, 1 |
| Wine | 50, {1: 1: 10}, 10 | 50, {1: 1: 10}, 10 | 50, {10}, 1 | 50, {10}, 1 |
| Contraceptive | 50, {1: 1: 10}, 10 | 50, {1: 1: 10}, 10 | 50, {10}, 1 | 50, {10}, 1 |
| Pima | 25, {1: 1: 10}, 10 | 25, {1: 1: 10}, 10 | 25, {10}, 1 | 25, {10}, 1 |
| Flare | 50, {1: 1: 10}, 10 | 50, {1: 1: 10}, 10 | 50, {10}, 1 | 50, {10}, 1 |
| ACA | 25, {1: 1: 10}, 10 | 25, {1: 1: 10}, 10 | 25, {10}, 1 | 25, {10}, 1 |

We compare with SCN, SCN+, IRVFL and IRVFL+ on five classification datasets and five regression datasets. The statistics of the datasets are illustrated in Table 1 and Table 2, including the number of training and test data, attributes, the number of the normal features, the number of the privileged features and classes or output. We randomly split attributes of each dataset mentioned above in half. All samples will be preprocessed with normalization.

The hyperparameters $C$ and $\gamma$ should be pre-defined by users. The settings of other parameters will be described in detail in Table 3. The user-defined parameters $C$ and $\gamma$ are chosen through random searches within $[10^{-2}, 10^{1}]$ and $[10^{2}, 10^{6}]$.

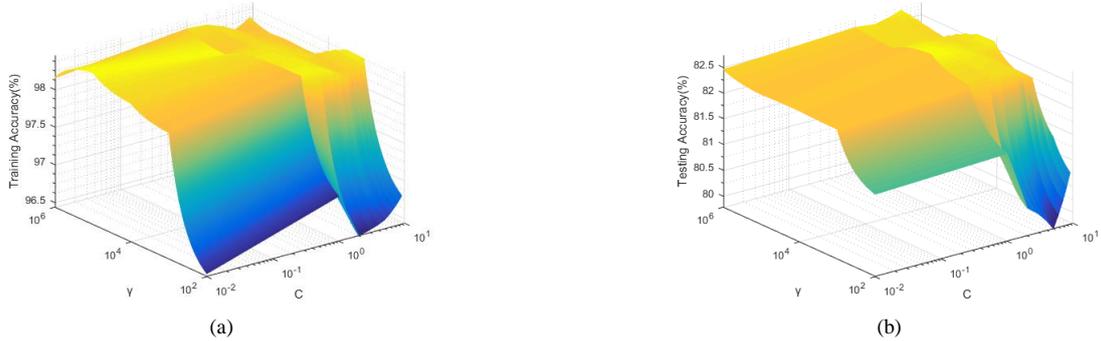

(a) (b)

Fig. 2. Performance of SCN+ with different user-defined parameters $C$ and $\gamma$ on Wine, where (a) and (b) represent training accuracy and testing accuracy respectively.

***Remark 2***：In order to make the graph easy to observe, all the data in Fig. 2 adopt the *log* function to normalize the data.

In theory, there is no exact theoretical basis to guide the selection of hyperparameters $C$ and $\gamma$, which can be only set empirically or repeat the experiment. Taking SCN+ as an example, the experiment is repeated on the Wine dataset, and each experiment $C$ and $\gamma$ are taken in the increasing sequences $[10^{-2}, 10^{-1}, 1, 2, 5, 10]$ and $[10^{2}, 10^{3}, \ldots, 10^{6}]$. The optimal parameters are selected according to the training and testing RMSE. Seeing from Fig. 2, we can intuitively see how to set $C$ and $\gamma$ can get the best solution. From (a), we can obtain that when $C$ is set to 1 and $\gamma$ is set to $10^{4}$, the training accuracy is the biggest, but (b) the test accuracy at this time is not, indicating that the generalization performance is not at its best. Considering (a) and (b), $C$ set as $10^{-1}$ and $\gamma$ as $10^{5}$ finally. In the following experiments, we will use this method to select the appropriate parameters $C$ and $\gamma$. Due to space limitations, the description of this process will be omitted later.

For classification datasets, all experiments will fix the maximum hidden layer nodes $L_{max}$ to compare the performance of each algorithm. As shown in Table 4, the training accuracy and test accuracy of SCN, SCN+, IRVFL and IRVFL+ are given. Besides, experiments on datasets Wine, Contraceptive, Pima, Flare and ACA fixed the maximum hidden layer nodes as 50, 50, 25, 50 and 25 respectively. The authors can obtain that the performance of SCN and SCN+ is better remarkably than that of IRVFL and IRVFL+, which means that the supervision mechanism of SCN can effectively improve the generalization of the model. More importantly, SCN+ can always achieve the best performance on all above classification datasets. Fig. 3 shows the training and testing accuracy curves of SCN, SCN+, IRVFL and IRVFL+ on different classification datasets. It should be noted that each node of each graph is the average of fifty times of training or testing. Like the results in Table 4, the training and testing accuracy of SCN and SCN+ is always better than IRVFL and IRVFL+. Although the training accuracy of SCN+ is not significantly better than that of SCN, the test accuracy has been improved relatively well. It should be noted that whether it is in Table 4 or Fig. 3, SCN+ and IRVFL+ can always perform better than SCN and IRVFL respectively. The above experimental results indicate that privileged information is effective in improving model performance on classification cases.

Table 4
Comparisons with three approaches on classification datasets in terms of training accuracy and testing accuracy.

| Datasets | Algorithms | Training Accuracy (%) | | Testing Accuracy (%) | |
|---|---|---|---|---|---|
| | | AVE | DEV | AVE | DEV |
| Wine $L_{max}=50$ | SCN | 98.36 | 0.66 | 82.54 | 1.46 |
| | **SCN+** | **98.38** | **0.45** | **82.74** | **1.16** |
| | IRVFL | 94.72 | 2.46 | 78.87 | 4.16 |
| | IRVFL+ | 95.04 | 1.94 | 80.08 | 4.92 |
| Contraceptive $L_{max}=50$ | SCN | 51.31 | 0.89 | 49.34 | 1.06 |
| | **SCN+** | **51.63** | **0.61** | **49.77** | **0.98** |
| | IRVFL | 50.09 | 1.62 | 47.28 | 2.23 |
| | IRVFL+ | 50.34 | 1.68 | 47.91 | 1.97 |
| Pima $L_{max}=25$ | SCN | 75.46 | 0.95 | 75.52 | 0.93 |
| | **SCN+** | **75.62** | **0.86** | **75.81** | **0.77** |
| | IRVFL | 66.66 | 1.27 | 70.51 | 1.55 |
| | IRVFL+ | 66.99 | 1.33 | 70.65 | 1.72 |
| Flare $L_{max}=50$ | SCN | 66.87 | 3.01 | 61.17 | 1.61 |
| | **SCN+** | **66.98** | **2.74** | **61.25** | **1.43** |
| | IRVFL | 61.85 | 4.53 | 59.05 | 2.90 |
| | IRVFL+ | 61.91 | 4.24 | 59.10 | 2.41 |
| ACA $L_{max}=25$ | SCN | 67.13 | 0.80 | 66.61 | **1.22** |
| | **SCN+** | **67.19** | **0.63** | **66.92** | 1.46 |
| | IRVFL | 66.15 | 1.86 | 64.28 | 2.39 |
| | IRVFL+ | 66.33 | 1.74 | 64.51 | 1.81 |

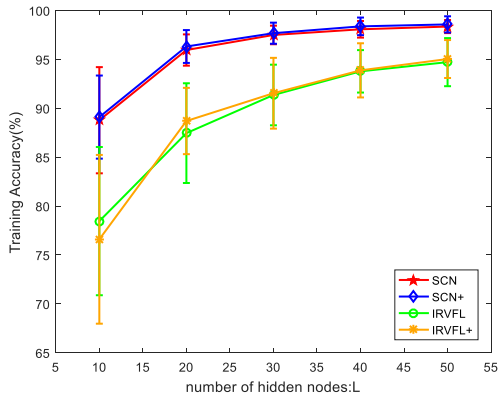

(a)

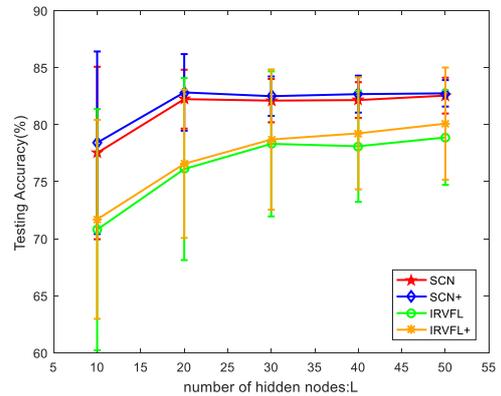

(b)

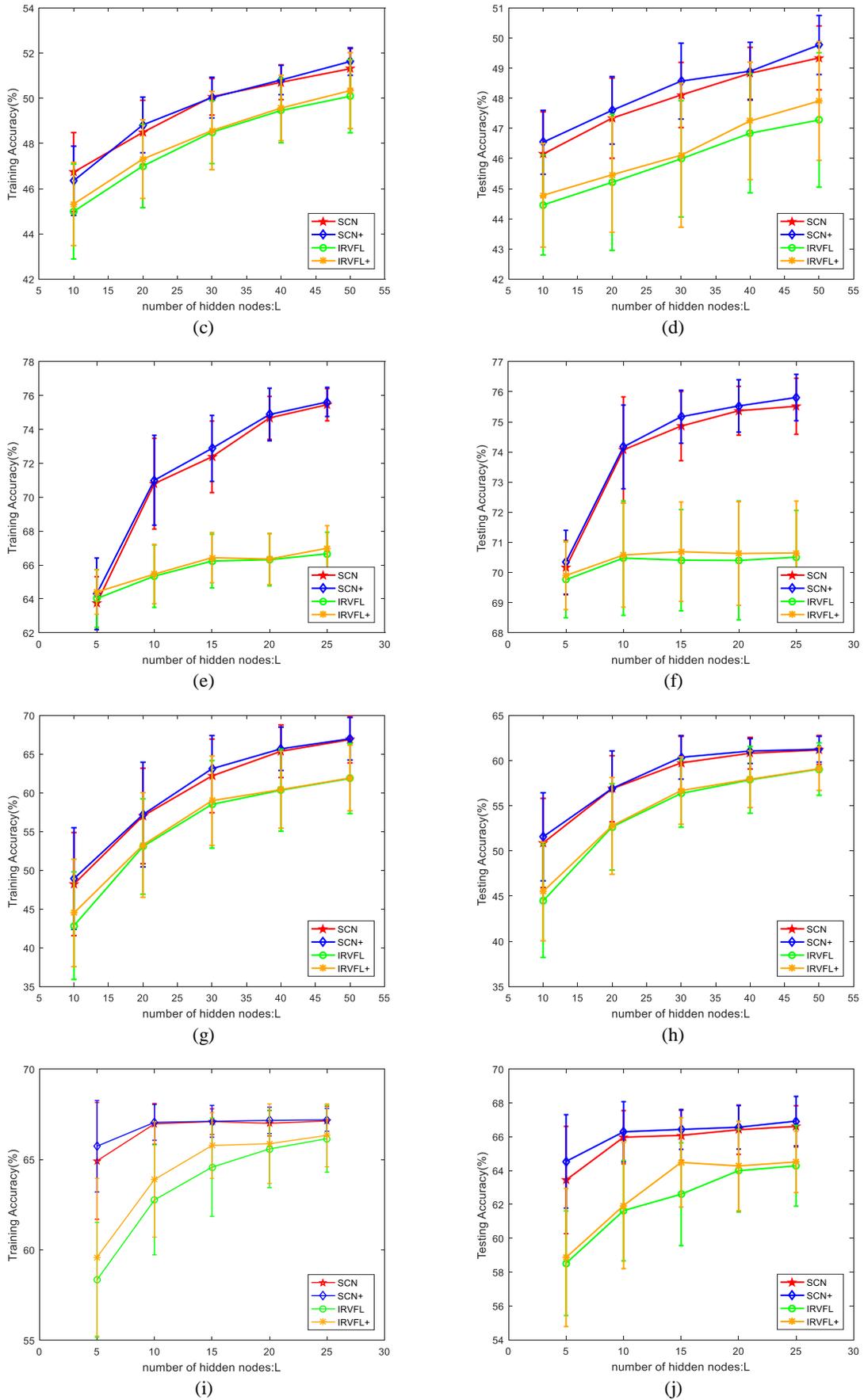

Fig. 3. Performance of SCN, SCN+ and IRVFL on classification datasets, where (a) and (b) represent the performance on Wine, (c) and (d) represent the performance on Contraceptive, (e) and (f) represent the performance on Pima, (g) and (h) represent the performance on Flare, (i) and (j) represent the performance on ACA.

For regression datasets, all experiments will fix the tolerance error ε to compare the performance of each algorithm. Table 5 shows the training RMSE, the test RMSE and the number of hidden layer nodes of each algorithm on different regression data sets under the condition of achieving the same expected error. The tolerance errors of datasets Mortgage, Treasury, ENB, Laser and W-Izmir are set to 0.1, 0.1, 0.2, 0.225 and 0.1 respectively. Expect on Mortgage the training RMSE of SCN perform better than SCN+, SCN+ can achieve best performance. More importantly, under the premise of achieving the same tolerance error, the hidden layer nodes required by SCN+ on Mortgage is 11.4% less than SCN, while IRVFL+ is 17.5% less than IRVFL; on Treasury, the hidden nodes number of SCN+ are 14.2% less than SCN, and IRVFL+ is 18.1% less than IRVFL; on ENB, the hidden nodes number of SCN+ are 4.0% less than SCN, and IRVFL+ is 4.6% less than IRVFL; on Laser, the hidden nodes number of SCN+ are 3.8% less than SCN, and IRVFL+ is 10.4% less than IRVFL; on W-Izmir, the hidden nodes number of SCN+ are 8.2% less than SCN, and IRVFL+ is 4.6% less than IRVFL. It can be seen that the SC algorithms can always get much smaller network structures than IRVFL and IRVFL+ under the condition of achieving the same tolerance error, which reduce the size of the models. Besides, SCN+ and IRVFL+ can get smaller model size than SCN and IRVFL respectively for adopting LUPI paradigm. Fig. 5 shows the curves of the training RMSE and testing RMSE of the four algorithms on different datasets. The authors can obtain that the SC algorithms always converges faster than IRVFL+ and IRVFL, except that SCN and IRVFL+ perform similarly on the ENB test set. Besides, under the premise of achieving the same desired solution, SCN+ requires fewer hidden nodes than SCN, while IRVFL+ requires more than SC algorithms and IRVFL requires the most. The above illustrates that SCN+ performs effectively on regression cases.

Table 5
Comparisons with three approaches on regression datasets in terms of training nodes($L$), training RMSE and testing RMSE.

| Datasets | Algorithms | Training RMSE | | Testing RMSE | | $L$ |
|---|---|---|---|---|---|---|
| | | AVE | AVE | AVE | DEV | |
| Mortgage ε=0.1 | SCN | **0.0962** | 0.0039 | 0.0956 | 0.0057 | 10.92 |
| | SCN+ | 0.0963 | **0.0034** | **0.0949** | 0.0056 | **9.68** |
| | IRVFL | 0.1005 | 0.0036 | 0.1020 | **0.0043** | 131.38 |
| | IRVFL+ | 0.0999 | 0.0021 | 0.1066 | 0.0074 | 108.04 |
| Treasury ε=0.1 | SCN | 0.0989 | **0.0012** | 0.1079 | 0.0034 | 19.08 |
| | SCN+ | **0.0987** | 0.0015 | **0.1068** | 0.0032 | **16.38** |
| | IRVFL | 0.1044 | 0.0137 | 0.1118 | 0.0102 | 147.80 |
| | IRVFL+ | 0.1008 | 0.0045 | 0.1133 | 0.0088 | 121.08 |
| ENB ε=0.2 | SCN | 0.1995 | 0.0011 | 0.2679 | 0.0288 | 24.86 |
| | SCN+ | **0.1992** | **0.0010** | **0.2676** | **0.0245** | **23.86** |
| | IRVFL | 0.2030 | 0.0075 | 0.3132 | 0.0554 | 73.30 |
| | IRVFL+ | 0.2023 | 0.0072 | 0.3120 | 0.0558 | 69.92 |
| Laser ε=0.225 | SCN | 0.2238 | **0.0011** | 0.2336 | 0.0020 | 20.26 |
| | SCN+ | **0.2237** | 0.0012 | **0.2335** | **0.0017** | **19.48** |
| | IRVFL | 0.2280 | 0.0075 | 0.2437 | 0.0073 | 66.78 |
| | IRVFL+ | 0.2247 | 0.0028 | 0.2434 | 0.0073 | 59.84 |
| W-Izmir ε=0.1 | SCN | 0.0964 | **0.0040** | 0.0998 | 0.0066 | 14.84 |
| | SCN+ | **0.0942** | 0.0047 | **0.0990** | **0.0061** | **13.62** |
| | IRVFL | 0.1048 | 0.0093 | 0.1072 | 0.0112 | 149.74 |
| | IRVFL+ | 0.1037 | 0.0080 | 0.1105 | 0.0116 | 142.90 |

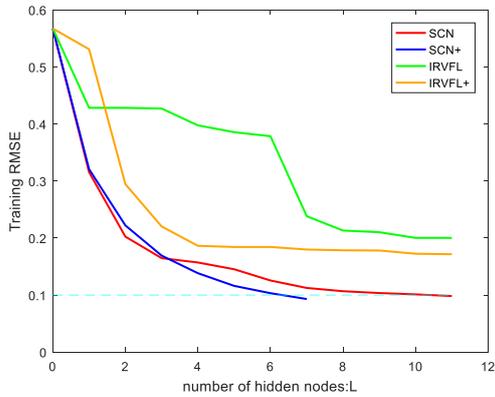

(a)

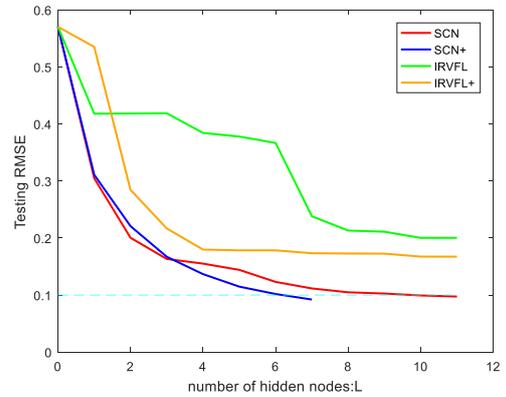

(b)

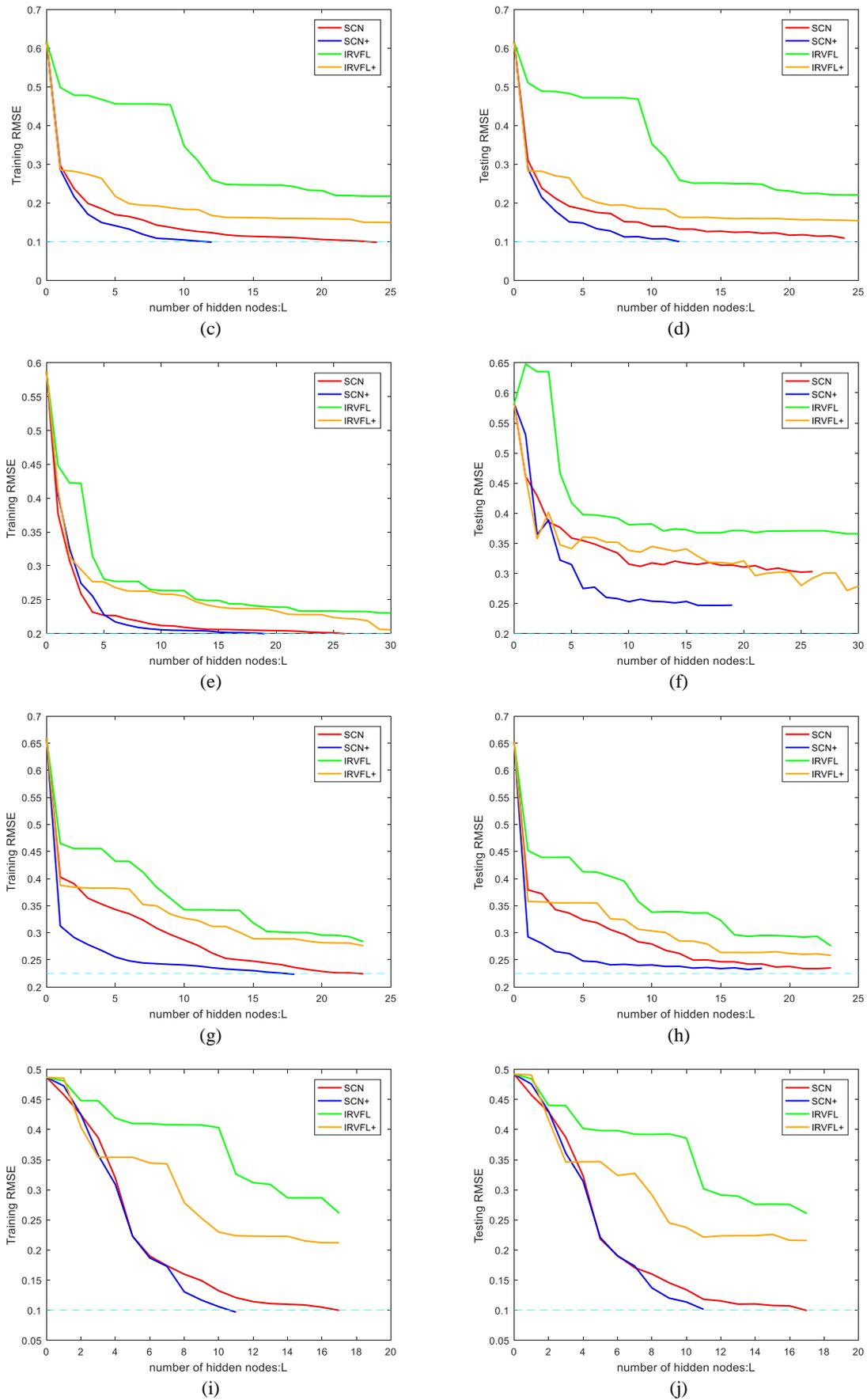

Fig. 4. Performance of SCN, SCN+ and IRVFL on regression datasets, where (a) and (b) represent the performance on Mortgage, (c) and (d) represent the performance on Treasury, (e) and (f) represent the performance on ENB, (g) and (h) represent the performance on Laser, (i) and (j) represent the performance on W-Izmir.

4.2. Discussion

Based on the above experimental results, we discuss the following points of the proposed method:

1) Obviously the hyperparameters $C$ and $\gamma$ have a great impact on the performance of SCN+. Although the hyperparameters set for different tasks are not exactly the same, they all have one thing in common: $C$ must be small enough, and $\gamma$ must be large enough. As mentioned above, $CH\beta$ and $C\tilde{H}_L\tilde{\beta}_L$ $CH_L\beta_L$ are slack functions which means the role of $C$ is a parameter to correct the error that should not be set too large. According to the ridge regression theory, as $\gamma$ increases, the variance of the model will be smaller, but $\beta$ will deviate from the true value that lead a larger model deviation. Therefore, $\gamma$ should be set large enough but not too large which needs to be determined according to the actual situation.

2) The authors analyzed two algorithms from the perspective of complexity [26]. The highest orders of SCN+ reaches $O(23N+4Nm+4)$ but SCN just $O(2N(m+1))$ for the number of data $N$, which indicates that SCN+ will consume more computing resources, especially on large-scale data. How to improve the computational efficiency of SCN+ is the focus of our future research.

## 5. Conclusions

By using LUPI paradigm, an incremental learner model based on SCN is proposed, named SCN+. SCN+ leverages privileged information into the IRVFL networks in the training stage. The convergence of SCN+ has been analyzed in this paper, which provides a strong theoretical guarantee. Through experiments on six real-world datasets, it is shown that the proposed SCN+ can effectively improve the generalization performance of the model. This newly-derived SCN+ can merge the incremental learning algorithm with the LUPI paradigm effectively, which provides a way of thinking for the application of LUPI paradigm in incremental learning.